\documentclass[letterpaper, 10 pt, journal, twoside]{IEEEtran}
\usepackage{amsmath,amsfonts}
\usepackage{algorithm}
\usepackage{algpseudocode}
\usepackage{array}
\usepackage[caption=false,font=normalsize,labelfont=sf,textfont=sf]{subfig}
\usepackage{textcomp}
\usepackage{stfloats}
\usepackage{url}
\usepackage{verbatim}
\usepackage{graphicx}
\usepackage{cite}
\usepackage{threeparttable}
\usepackage{multirow}
\usepackage{dsfont}
\usepackage{makecell}
\usepackage{booktabs}
\usepackage{soul, color, xcolor}
\soulregister\cite7
\soulregister\ref7 
\definecolor{myred}{RGB}{192, 0, 0}
\definecolor{myblue}{RGB}{0, 112, 192}
\definecolor{mypurple}{RGB}{120,32,110}
\usepackage{hyperref}
\newcommand{\highlight}[1]{\textcolor{black}{#1}}

\begin{document}

\title{
Visual-Tactile Peg-in-Hole Assembly Learning from \\Peg-out-of-Hole Disassembly
}

\author{Yongqiang Zhao$^1$, Xuyang Zhang$^1$, Zhuo Chen$^1$, Matteo Leonetti$^2$, \\Emmanouil Spyrakos-Papastavridis$^1$, Shan Luo$^1$ 
\thanks{Manuscript received: December 4, 2025; Revised February 22, 2026; Accepted March 16, 2026.}
\thanks{This paper was recommended for publication by Editor Ashis Banerjee upon evaluation of the Associate Editor and Reviewers’ comments. This work was supported in part by the EPSRC projects ``ViTac: Visual-Tactile Synergy for Handling Flexible Materials" (EP/T033517/2) and "TacDiff: Designing Tactile-based Robots via Differentiable Simulations". (Corresponding author: Shan Luo.)}
\thanks{$^1$Yongqiang Zhao, Xuyang Zhang, Zhuo Chen, Emmanouil Spyrakos-Papastavridis, and Shan Luo are with Department of Engineering, King’s College London, Strand, London,
WC2R 2LS, United Kingdom, \{yongqiang.zhao, xuyang.zhang, zhuo.7.chen, emmanouil.spyrakos, shan.luo\}@kcl.ac.uk.}
\thanks{ $^2$Matteo Leonetti is with Department of Informatics, King’s College London, Strand, London,
WC2R 2LS, United Kingdom, matteo.leonetti@kcl.ac.uk.
}
\thanks{Digital Object Identifier (DOI): see top of this page.}
}

\markboth{IEEE ROBOTICS AND AUTOMATION LETTERS. PREPRINT VERSION. ACCEPTED March, 2026}%
{Zhao \MakeLowercase{\textit{et al.}}: Visual-Tactile Peg-in-Hole Assembly Learning from Peg-out-of-Hole Disassembly}

\maketitle
\pagestyle{headings}

\begin{abstract}
Peg-in-hole (PiH) assembly is a fundamental yet challenging robotic manipulation task. While reinforcement learning (RL) has shown promise in tackling such tasks, it requires extensive exploration. In this paper, we propose a novel visual-tactile skill learning framework for the PiH task that leverages its inverse task, i.e., peg-out-of-hole (PooH) disassembly, to facilitate PiH learning. Compared to PiH, PooH is inherently easier as it only needs to overcome existing friction without precise alignment, making data collection more efficient. To this end, we formulate both PooH and PiH as Partially Observable Markov Decision Processes (POMDPs) in a unified environment with shared visual-tactile observation space. A visual–tactile PooH policy is first trained; its trajectories, containing kinematic, visual and tactile information, are temporally reversed and action-randomized to provide expert data for PiH. In the policy learning, visual sensing facilitates the peg–hole approach, while tactile measurements compensate for peg–hole misalignment. Experiments across diverse peg–hole geometries show that the visual–tactile policy attains 6.4\% lower contact forces than its single-modality counterparts, and that our framework achieves average success rates of 87.5\% on seen objects and 77.1\% on unseen objects, outperforming direct RL methods that train PiH policies from scratch by 18.1\% in success rate.
Demos, code, and datasets are available at \href{https://sites.google.com/view/pooh2pih}{https://sites.google.com/view/pooh2pih}.
\end{abstract}


\section{INTRODUCTION}
\IEEEPARstart{T}he peg-in-hole (PiH) task presents a significant challenge in robotic assembly and plays a crucial role in both industrial applications and everyday activities such as gear assembly and USB insertion~\cite{chen2023multimodality}. Research on this problem is fundamental to robotics, as it contributes to advancing general assembly capabilities and improving robot dexterity. Due to its importance, the PiH problem has been one of the most active research areas in robotics in recent years. It can be formulated as a target localization problem \cite{dong2021tactile}, which requires precise insertion of a peg into a hole. 

Despite extensive research~\cite{chen2023multimodality, dong2021tactile}, PiH remains challenging due to the large amount of exploration required. Reinforcement learning (RL) methods, which train policies through trial and error, are robust to target location variations ~\cite{dong2021tactile, zhao2023skill, bi2026alore} but demand substantial data and interaction. Supervised learning (SL) from expert demonstrations ~\cite{zhao2026vitac} can accelerate training and achieve high success rates, yet relies on high-quality demonstrations and often generalizes poorly to unseen scenes. RL with prior knowledge ~\cite{tang2024automate} alleviates some of these issues, but collecting such priors on PiH still requires careful demonstrations and intensive peg–hole interaction, causing time cost and hardware wear.

\begin{figure}
	\centering
	\includegraphics[scale=0.5]{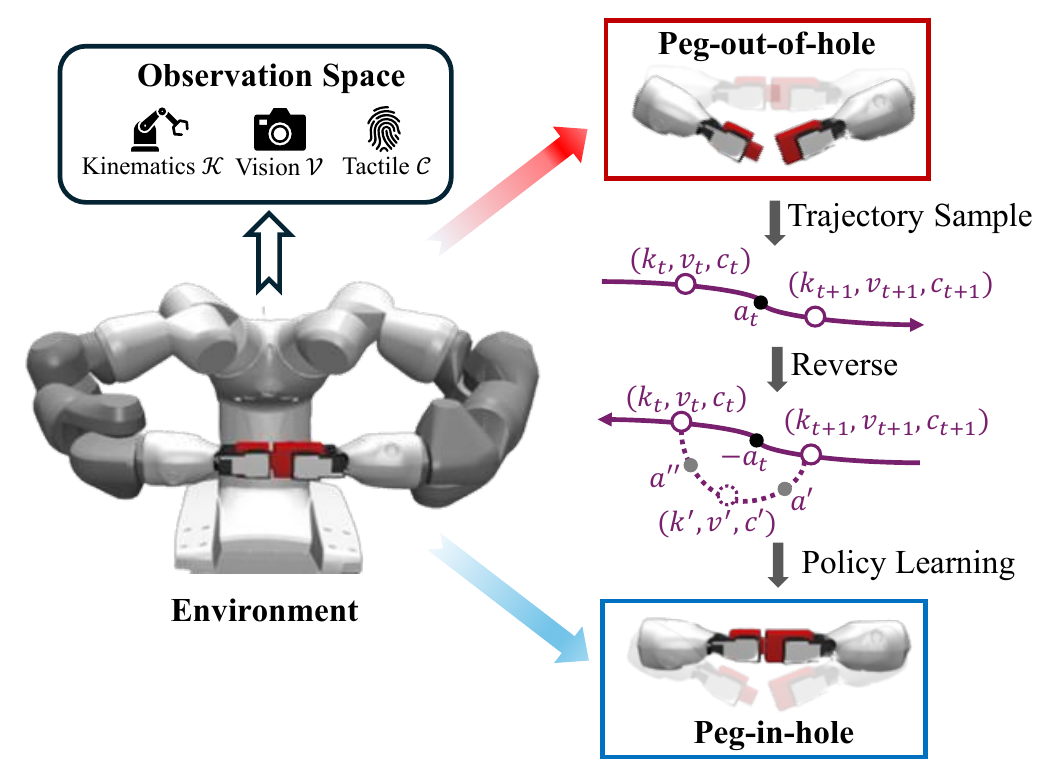}
        \caption{Key concept of peg-in-hole (PiH) skill learning from a peg-out-of-hole (PooH) policy. Both tasks share the same visual–tactile observation space $(k_t, v_t,c_t) \in (\mathcal{K},\mathcal{V},\mathcal{C})$ and action space $a_t$. A PooH policy is first learned with reinforcement learning, and its trajectories are temporally reversed to generate PiH training data. Randomized actions $a'$ and $a''$ (dashed curve) further enrich contact patterns and reduce the contact gap between PooH and PiH.}
        \label{fig.overview}
	\vspace{-10pt}
\end{figure}

Research in human neurology shows that observing or practicing the reverse of a manipulation task activates complementary action circuits and supports bidirectional action–effect learning~\cite{hommel2019theory}. Robotics studies have similarly leveraged disassembly to facilitate assembly, for example assembling LEGO structures after first breaking them apart ~\cite{walsman2022break}. In PiH, insertion typically demands precise force control and compliant gripping, whereas peg-out-of-hole (PooH) can often be achieved by simpler pulling actions~\cite{goli2024jamming}. Despite this asymmetry, PooH can provide informative contact experiences that simplify PiH exploration. Motivated by this, we propose to learn a visual–tactile PooH policy and reverse its trajectories to generate informative training data for PiH insertion, in which both tasks share the same visual–tactile observation space and action space.

Sensing modality is another key factor in assembly. Vision supports object recognition and coarse pose estimation~\cite{chen2024vision, ou2024hybrid}, while tactile/force sensing helps detect contact, jamming, and misalignment~\cite{dong2021tactile}. Single-modality methods suffer from modality-specific limitations: vision-only approaches are sensitive to occlusion and cannot directly sense forces~\cite{huang2017vision}, whereas tactile-only approaches lack long-range guidance to the hole~\cite{zhao2024fots}. Following~\cite{tang2025visual}, we employ multi-modal sensing to exploit complementary strengths. By collecting visual–tactile trajectories on the simpler PooH task, we obtain rich multi-modal demonstrations that can be reused for the harder PiH insertion without additional risky exploration.

In this work, we propose a visual–tactile skill learning framework for PiH assembly. We formulate both PooH and PiH as Partially Observable Markov Decision Processes (POMDPs) in a unified environment with visual and tactile observations. A PooH policy is first trained with RL, and its trajectories are inverted to construct training data for PiH policy learning. During this process, we introduce action randomization to enrich contact configurations and bridge the distinct force characteristics between PiH and PooH. The proposed framework substantially improves success rates over direct RL that trains PiH policies from scratch. Extensive experiments on diverse peg–hole geometries demonstrate strong generalization and low contact force, achieving average success rates of 87.5\% on seen objects and 77.1\% on unseen objects

In summary, the contributions of this work are as follows:
\begin{itemize}
    \item We propose a novel PooH-to-PiH skill learning framework that formulates both PooH and PiH as POMDPs, enabling robots to efficiently acquire insertion skills from the inverse task;
    \item Visual and tactile sensing are integrated to facilitate peg-hole approaching and correct peg-hole misalignment, respectively, jointly enhancing PiH performance; 
    \item Extensive experiments were conducted that show the effectiveness and generalizability of our method, compared with the direct RL methods.
\end{itemize}

\section{RELATED WORKS}
\subsection{Robot Peg-in-hole Assembly}
Solutions to the peg-in-hole (PiH) problem depend on sensing, robot hardware, and the initial peg/hole poses. Recent work mainly follows two paradigms: reinforcement learning (RL) and supervised learning (SL). Learning PiH skills with RL is common in recent studies. Dong et al. \cite{dong2021tactile} and Zhao et al.~\cite{zhao2024fots} used off-policy RL with tactile images from vision-based tactile sensors~\cite{gomes2020geltip}, while Chen et al. ~\cite{chen2023multimodality} addressed multi-object PiH by combining Soft Actor-Critic (SAC) with Hindsight Experience Replay (HER). However, continuous action spaces and high-dimensional, contact-rich states make PiH exploration-intensive and slow to train.

Supervised learning methods map sensor observations to actions using expert data. Spector et al.~\cite{spector2021insertionnet} formulated PiH as a regression problem and learned a residual policy from heavily augmented demonstrations. Although SL is generally more sample-efficient than RL, it requires large labeled datasets and often generalizes poorly to unseen objects, clearances, or poses.

Some works combine the strengths of RL and SL by injecting prior knowledge into RL. Tang et al.~\cite{tang2024automate} trained a PiH policy with an imitation objective, where assembly demonstrations are obtained via disassembly (i.e., grasp sampling). However, this data collection simply reverses the kinematic paths without accounting for the distinct contact configurations between PiH and PooH, where the PiH process involves more adjustment than the PooH process. Nguyen et al.~\cite{nguyen2024symmetry} improve PiH sample efficiency by exploiting object symmetries via data augmentation and symmetry-consistency losses under partial observability, which typically assumes a known symmetry group and focuses on direct PiH learning. In contrast, we derive PiH data from PooH and introduce action randomization to diversify the contact configurations in the reversed PooH trajectories, without relying on object symmetry.

\subsection{Sensing Modalities for Robot Assembly}
Robotic assembly relies on rich sensory information from the robot, objects, and environment~\cite{xu2019compare}. Proprioception provides the robot’s internal state, while force/torque, tactile, and visual sensing capture external interactions.
Force/torque sensing is widely used in PiH, as it enables estimation of contact configurations. Ortega et al.~\cite{ortega2021dual} employed dual force/torque sensors to improve the success rate. Tactile sensing provides an alternative with richer contact information. Vision-based tactile sensors~\cite{gomes2020geltip} provide high-resolution feedback and are widely applied to PiH. Dong et al.~\cite{dong2021tactile} and Zhao et al.~\cite{zhao2024fots} exploited tactile flow to infer contact configurations and refine actions. However, these force/torque-based and tactile-based methods mainly support the local adjustment phase and are ineffective during the approach phase.

Visual perception is critical for object recognition and pose estimation. Chen et al.~\cite{chen2024vision} proposed a vision-based algorithm for general PiH, Huang et al.~\cite{huang2017vision} used vision to correct positioning errors on a Baxter robot, and Sileo et al.~\cite{sileo2024vision} adopted a CNN to improve hole localization. Nonetheless, vision-only methods are sensitive to occlusion and cannot directly sense jamming or excessive forces.

To overcome these limitations, multi-modal sensing has been explored. Li et al.~\cite{li2024end} combined visual and force cues for force-constrained assembly, and Tang et al.~\cite{tang2025visual} proposed a visual–tactile fusion scheme for dynamic environments. Motivated by these complementary strengths, we model kinematic, visual, and tactile signals as a unified observation space for both PooH and PiH POMDPs. This allows multi-modal trajectories collected in the easier PooH task to be reused to accelerate learning robust PiH insertion policies.

\begin{figure*}[ht]
	\centering
	\includegraphics[scale=0.45]{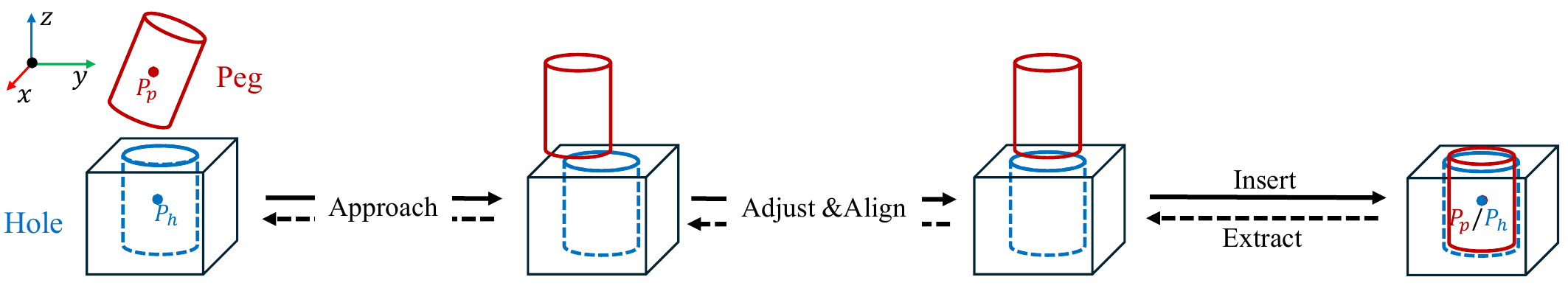}
        \caption{The process of peg-in-hole (solid line) and peg-out-of-hole (dashed line). For peg-in-hole process, the peg approaches the hole, then inserts into the hole with adjusting and aligning. For peg-out-of-hole, the process is the opposite, but with rare adjusting and aligning.}
        \label{fig.problem}
	\vspace{-10pt}
\end{figure*}

\section{PROBLEM STATEMENT}
\label{SEC:3}
The peg-in-hole (PiH) task is defined as planning and executing a motion that enables a robot to insert a rigid peg into a matching hole without applying excessive force. As shown in Fig.~\ref{fig.problem}, this typically involves four stages: approach, adjust, align, and insert~\cite{lee2022peg}. In contrast, the peg-out-of-hole (PooH) task replaces insertion with extraction and usually requires little or no adjusting \& aligning. Ignoring frictional disturbances and minor corrections, the peg–hole poses in PooH approximately follow the time-reversed trajectories of PiH in both the kinematic and visual spaces.

Let $\mathcal{P}_p$ and $\mathcal{P}_h$ denote the peg and hole poses in Cartesian space. During insertion ($t=0 \rightarrow T$), the peg moves from a distant pose toward the hole until they coincide, i.e., $\mathcal{P}_p(T)\approx\mathcal{P}_h(T)$. During extraction ($t'=0 \rightarrow T'$), the motion starts from a coincident configuration and moves away. This establishes an approximate bijection between PiH and PooH trajectories in kinematic and visual spaces, enabling us to reuse PooH rollouts for PiH by temporal reversal. However, the tactile space is not reversible: insertion is contact-rich and involves jamming and incremental alignment, whereas extraction can remain nearly contact-free once the peg is released. Simply reversing PooH trajectories therefore fails to reproduce the diverse contact patterns required for robust PiH insertion. To bridge this tactile asymmetry, we introduce action randomization along PooH trajectories after reversal, enriching contact configurations and making the induced tactile signals more representative of those encountered in PiH.

We address both PooH and PiH with the Soft Actor-Critic (SAC) algorithm~\cite{haarnoja2018soft} in a shared robotic environment with identical robot configuration, sensors, and action interfaces. Since the peg-hole poses $\mathcal{P}=(\mathcal{P}_p,\mathcal{P}_h)$ are not directly observable in real-world settings, 
we infer it from sensory feedback and formulate both tasks as Partially Observable Markov Decision Processes (POMDPs) defined by the tuple
$\left(\mathcal{S}, \mathcal{A}, \mathcal{O}, p, \Omega, r, \gamma\right)$,
where $\mathcal{S}$ denotes the state space and $\mathcal{A}$ is a continuous action space. The observation space is $\mathcal{O} = (\mathcal{K}, \mathcal{V}, \mathcal{C})$, with $\mathcal{K}$, $\mathcal{V}$, and $\mathcal{C}$ representing kinematic, visual, and tactile observations, respectively. The state transition dynamics are given by $p(s_{t+1}\mid s_t, a_t)$. $\Omega(o_{t+1}\mid s_{t+1}, a_t)$ represents the observation model, where $o_t \in \mathcal{O}$. The reward function is $r:\mathcal{S}\times\mathcal{A}\rightarrow\mathbb{R}$.
Finally, $\gamma\in[0,1)$ denotes the discount factor.
Because PooH and PiH share the same multi-modal observation and action spaces, the randomized and reversed PooH trajectories naturally serve as expert-like data for PiH policy learning, reducing the need for extensive exploration and long training times.
Details of the action and observation design, as well as the reward function, are provided in the next section.

\section{METHODS}
\label{SEC:4}

\begin{figure*}[ht]
	\centering
	\includegraphics[scale=0.52]{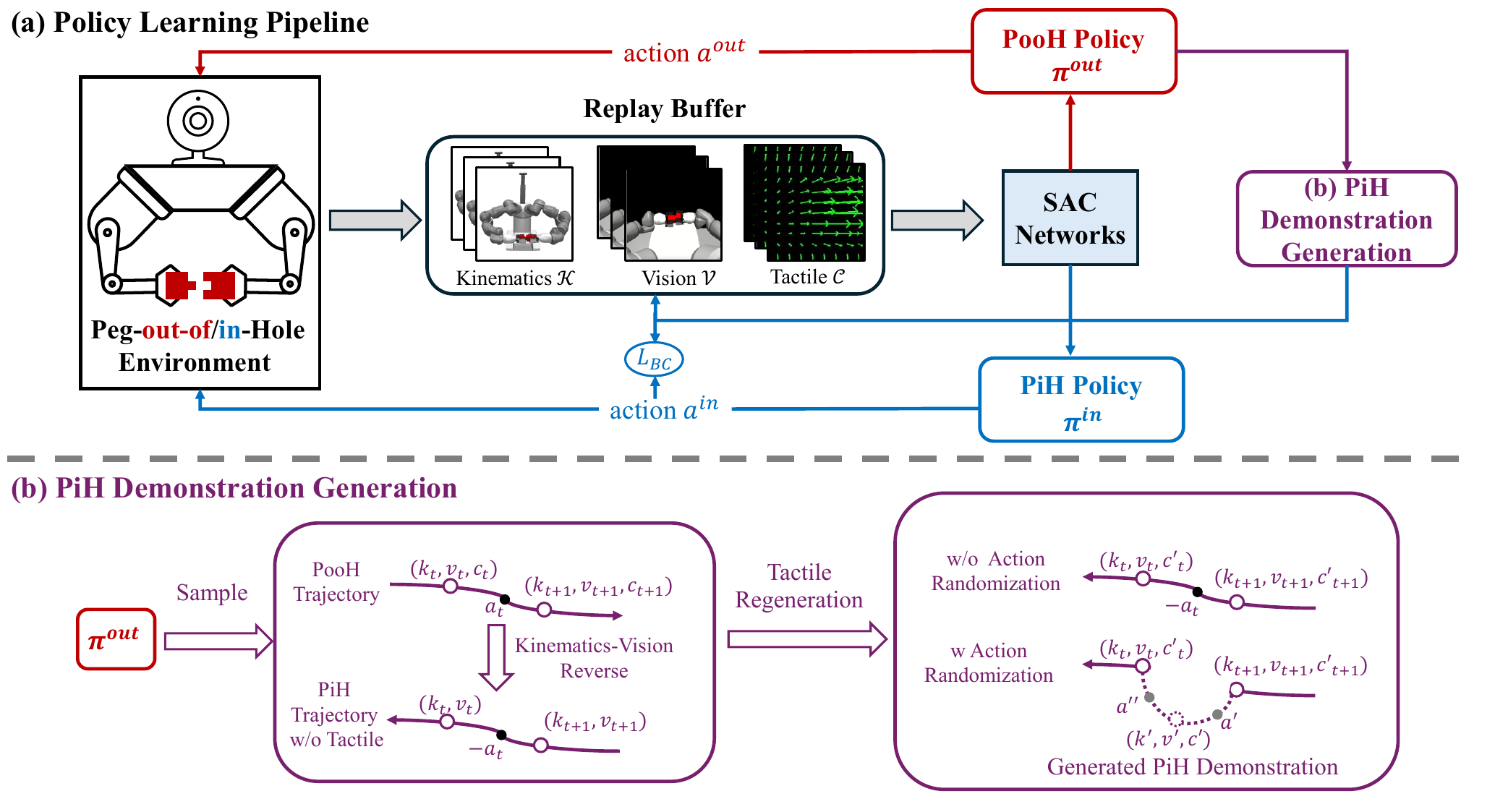}
        \caption{(a) Pipeline of the proposed peg-in-hole (PiH) skill learning framework. We construct a common environment for peg-out-of-hole (PooH) and PiH tasks, with robot kinematics, tactile flows and visual images as inputs. PooH policy is trained using Soft Actor-Critic (SAC) firstly. Then, we generate PiH demonstrations based on the trained PooH policy. Finally, we train the PiH policy via adopting hybrid replay buffer and behavior cloning loss. \textcolor{myred}{Red lines} represent the PooH policy learning, \textcolor{mypurple}{purple lines} represent the PiH demonstration generation, and \textcolor{myblue}{blue lines} represent the PiH policy learning. (b) The details of PiH demonstration generation, where we firstly reverse the PooH trajectories to generate PiH demonstrations without tactile observation, and then regenerate the tactile information combined with action randomization.}
        \label{fig.framework}
	\vspace{-10pt}
\end{figure*}

As Fig. \ref{fig.framework} shows, the proposed visual-tactile skill-learning framework comprises three stages: (1) peg-out-of-hole (PooH) learning, (2) peg-in-hole (PiH) demonstration generation, and (3) peg-in-hole policy learning. First, the PooH skill is acquired through the Soft Actor-Critic (SAC) algorithm with kinematic, visual and tactile observations. Second, a large dataset of PooH trajectories is generated, and then these trajectories are reversed to create PiH data combined with action randomization. Finally, these PiH data are used to train a PiH policy in the same environment as the PooH task.

\subsection{A Unified Visual-Tactile Agent for Peg-out-of/in-hole}
\label{SEC:4B}
In this study, we construct a unified manipulation environment for both the PooH and PiH tasks.

\textbf{Observations:} The agent’s observation space consists of three categories of sensory inputs.

\begin{itemize}
\item \textbf{Robot kinematics:} it provides the Cartesian pose of the robot end-effectors. Each end-effector's state is represented as a six-dimensional vector
$k_t = (x_t, y_t, z_t, \theta_{xt}, \theta_{yt}, \theta_{zt}) \in \mathcal{K}$,
corresponding to the translational and rotational components of the end-effector pose.
\item \textbf{Visual:} it provides global contextual information through an image observation $v_t \in \mathcal{V}$ captured by a camera with a resolution of $96\times96$ pixels. 
\item \textbf{Tactile:} it provides local contact information derived from tactile marker flow in vision-based tactile sensors. Following~\cite{xue2025reactive}, the raw deformation field is processed using Principal Component Analysis (PCA) and reduced to a 15-dimensional feature vector $c_t \in \mathcal{C}$.
\end{itemize}

\textbf{Actions:}
Since the gripper operates freely in Cartesian space, each action corresponds to a target displacement of the end-effector, defined as
\begin{equation}
 a_t=k_{t+1}-k_t,
\end{equation}
representing position and orientation changes between consecutive time steps.

\textbf{Reward:} Both PooH and PiH tasks can be regarded as goal-conditioned tasks, hence we design the reward function $\mathcal{R}$ as the negative of the distance between the desired goals and objects.
\begin{equation}
\label{eq:1}
    \mathcal{R}=-||\mathcal{P}(t)-\mathcal{G}||_2^2
\end{equation}
where $\mathcal{P}(t)=(\mathcal{P}_p(t), \mathcal{P}_h(t))$ and $\mathcal{G}=(\mathcal{G}_p, \mathcal{G}_h)$, $\mathcal{G}_p$ is the desired goal of the peg, while $\mathcal{G}_h$ represents the desired goal of the hole. All the object and goal poses are privileged and accessible in simulation. The differences between PooH and PiH tasks lie in the initial object poses and the desired goals.

\subsection{Peg-out-of-hole Policy Learning}
We train the PooH policy $\pi^{out}$ in simulation using reinforcement learning with multi-modal observations $o_t=(k_t, v_t, c_t)$, where $k_t$, $v_t$, and $c_t$ denote robot kinematics, visual, and tactile inputs, respectively. A PooH trajectory is given by
$\tau^{out} = \{(o_t, a_t, o_{t+1})\}_{t=0}^{T-1}$,
with actions selected as $a_t = \pi^{out}(o_t)$.

Because the task requires continuous control, we adopt the Soft Actor-Critic (SAC) algorithm~\cite{raffin2021stable} to learn $\pi^{out}$ in an off-policy, sample-efficient manner. SAC maximizes a maximum-entropy objective,
\begin{equation}
J(\pi^{out}) = \sum_{t=0}^{T-1} \mathbb{E}_{(o_t, a_t) \sim \rho_{\pi^{out}}}
\left[ r_t + \alpha \mathcal{H}\big(\pi^{out}(\cdot \mid o_t)\big) \right],
\end{equation}
where $r_t$ is the task reward, $\rho_{\pi^{out}}$ is the observation–action distribution under $\pi^{out}$, $\mathcal{H}$ is the policy entropy, and $\alpha$ controls the entropy weight.

\textbf{Policy training.} All observations are encoded before being fed to the SAC actor–critic. Robot kinematics and tactile signals are concatenated and passed through an MLP, while visual inputs are processed by a CNN; the resulting features are fused and used as input to the policy and value networks. To improve robustness and generalization, we randomize initial object poses within a sub-region of the robot workspace for each episode. Training a visual–tactile PooH policy ensures that each collected trajectory captures both long-range visual approach cues and local contact information, which are later reused as PiH demonstrations after temporal inversion with action randomization.

\subsection{Peg-in-hole Demonstration Generation}
\label{SEC:3C}
Once the PooH policy $\pi^{out}$ is trained, it can be used to replay the PooH task to collect a set of PooH trajectories $\tau^{out}$: 
\begin{equation}
\begin{aligned}
    &\tau^{out} = \cdots \rightarrow(o_t,a_t,r^{out}_t, o_{t+1})\rightarrow\cdots\\
    =&\cdots\rightarrow((k_t,v_t,c_t), a_t, r^{out}_t, (k_{t+1},v_{t+1},c_{t+1})) \rightarrow \cdots,
\end{aligned}
\end{equation}
where $r^{out}_t$ is the reward in the PooH task.

Due to the dynamic gap between the PooH and PiH tasks, we first reverse the PooH process in the kinematic–visual space, ignoring tactile observations. The robot executes inverse actions so that the original PooH trajectory becomes an approximate PiH trajectory $\hat{\tau}^{in}$ without tactile input:
\begin{equation}
\begin{aligned}
    \hat{\tau}^{in} &=\cdots \rightarrow ((k_{t+1},v_{t+1}),  -a_t, r^{in}_t, (k_t, v_t)) \rightarrow \cdots,
\end{aligned}
\end{equation}
where $r^{in}_t \in \mathcal{R}$ is calculated based on the desired goal of the peg-in-hole task.

After inverting kinematic and visual data, we replay the reversed kinematic trajectory in simulation and regenerate tactile observations at each timestep using the tactile simulator based on simulated contacts, yielding complete visual--tactile observations.
As discussed in Section~\ref{SEC:3}, PooH and PiH are only approximately time-symmetric: the approach and axis-aligned extraction in PooH roughly reverse to PiH, but insertion-specific alignment/jamming-recovery does not.
To better align the reversed PooH data with the PiH task, we add action randomization during trajectory reversal (Algorithm~\ref{alg:pooh2pih}). Randomization is activated when the peg-hole relative $z$ distance falls below $d=0.01$m. We sample $a'=[\Delta x,\Delta y,d]$ with $\Delta x,\Delta y\in[-0.02,0.02]$\,m to ensure contact, increasing contact diversity and capturing more realistic insertion dynamics.
Hence, the PiH trajectory with action randomization is:
\begin{equation}
\begin{aligned}
\cdots\rightarrow
&\big((k_{t+1},v_{t+1},c'_{t+1}), a', r^{in'}, (k',v',c')\big)\\
&\rightarrow\big((k',v',c'), a'', r^{in''}, (k_t,v_t,c'_t)\big)\rightarrow\cdots,
\end{aligned}
\end{equation}
where $-a_t=a'+a''$, $a'$ is the randomized action, and $o'=(k',v',c')$ is the intermediate observation after executing $a'$. The recovery action $a''$ returns the agent toward the original next state, with rewards $r^{in'}$ and $r^{in''}$. This yields richer contact configurations~\cite{dong2021tactile}. In our experiments, we apply action randomization to 50\% of the generated PiH trajectories.


\begin{algorithm}[t]
\caption{Generate PiH trajectory from PooH trajectory}
\label{alg:pooh2pih}
\begin{algorithmic}[1]
\Require PooH trajectory $\tau^{out}$; peg pose $P_p$; hole pose $P_h$; threshold $d=0.01\,\mathrm{m}$;
        $\Delta x,\Delta y \in [-0.02,0.02]\,\mathrm{m}$; $\texttt{count}=0$
\Ensure generated PiH trajectory $\tau^{in}$
\State $\tau^{in} \gets \emptyset$
\For{$t \gets T-1$ \textbf{downto} $0$}
  \State $\bigl((k_t,v_t),a_t,(k_{t+1},v_{t+1})\bigr) \gets \tau^{out}_t$
    \If{$\bigl\lvert (P_{p}(t)-P_{h}(t)) \bigr\rvert_{z} < d \ \textbf{and}\ \texttt{count}=0$}
        \State Sample $\Delta x, \Delta y$
        \State $a' \gets [\Delta x,\Delta y,d]$, \quad $k' \gets k_{t+1}+a'$
        \State Generate visual and tactile information $v',c'$
        \State $\tau^{in} \gets \bigl((k_{t+1},v_{t+1},c'_{t+1}),\, a',\, (k',v',c')\bigr)$
        \State $a'' \gets -a_t - a'$
        \State Generate tactile information $c'_t$
        \State $\tau^{in} \gets \bigl((k',v',c'),\, a'',\, (k_t,v_t,c')\bigr)$
        \State $\texttt{count} \gets 1$
    \Else
        \State Generate tactile information $c'_t$
        \State $\tau^{in} \gets \bigl((k_{t+1},v_{t+1},c'_{t+1}),\, -a_t,\, (k_t,v_t,c'_t)\bigr)$
    \EndIf
\EndFor
\State \Return $\tau^{in}$
\end{algorithmic}
\end{algorithm}

\subsection{Peg-in-Hole Policy Learning}
\label{SEC:4D}

Using the generated demonstrations for the peg-in-hole (PiH) task, we train the PiH policy $\pi^{in}$ within the same environment as the peg-out-of-hole (PooH) task. The two tasks share identical state, observation, and action spaces, and the reward function retains the same structure. We employ the off-policy Soft Actor-Critic (SAC) algorithm to train $\pi^{in}$ and incorporate two auxiliary modules to accelerate training and enhance performance.

\textbf{Hybrid Replay Buffers:}
Off-policy reinforcement learning methods, such as SAC, maintain a replay buffer for sampling transition tuples $(o_t, a_t, r_t, o_{t+1})$. In addition to the standard buffer that stores data collected through environment interaction, we introduce an expert replay buffer containing transitions from the generated PiH trajectories. During training, samples are drawn jointly from both buffers in a hybrid manner, i.e., each minibatch is formed by mixing transitions from the standard and expert replay buffers according to a sample ratio. An annealing mechanism gradually decreases the sampling ratio from the expert replay buffer (from 0.3 to 0.0), allowing the policy to transition from imitation-based to fully autonomous learning as training progresses.

\textbf{Behavior Cloning Loss:}
To further exploit the generated expert data, we incorporate a behavior cloning (BC) loss during RL training, using the PiH demonstrations as ground truth. In addition to standard RL updates, the policy performs several auxiliary BC updates after each rollout by maximizing the log-likelihood of expert actions under the current policy. This guides the actor toward expert-like behavior, improving both efficiency and stability. The BC loss is defined as:
\begin{equation}
\mathcal{L}_{\text{BC}} = - \lambda \mathbb{E}_{(o^,a) \sim \mathcal{D}^{in}} \left[ \log \pi^{in}_{\theta}(a \mid o) \right],
\end{equation}
where $\lambda$ is a weighting coefficient, $\mathcal{D}^{in}$ denotes the distribution of the generated PiH data, and $\pi^{in}_{\theta}(a \mid o)$ represents the actor’s probability of producing action $a$ in a visual-tactile observation $o$. An annealing schedule is applied to $\lambda$, decreasing it from 0.05 to 0.0 to prevent excessive reliance on imitation during the later stages of training.

\textbf{PiH Policy Training:} First, the visual image $\mathcal{V}$ is fed into a visual encoder, composed of Convolutional Neural Networks (CNNs), while the robot kinematic information $\mathcal{K}$ and tactile information $\mathcal{C}$ are concatenated and then fed into an MLP. The extracted features $z^{in}$ are then passed through the SAC network to predict the action $a$. Once training is complete, we can obtain a PiH policy $\pi^{in}$ trained based on the PooH policy.

\begin{figure}[!t]
	\centering
	\includegraphics[scale=0.35]{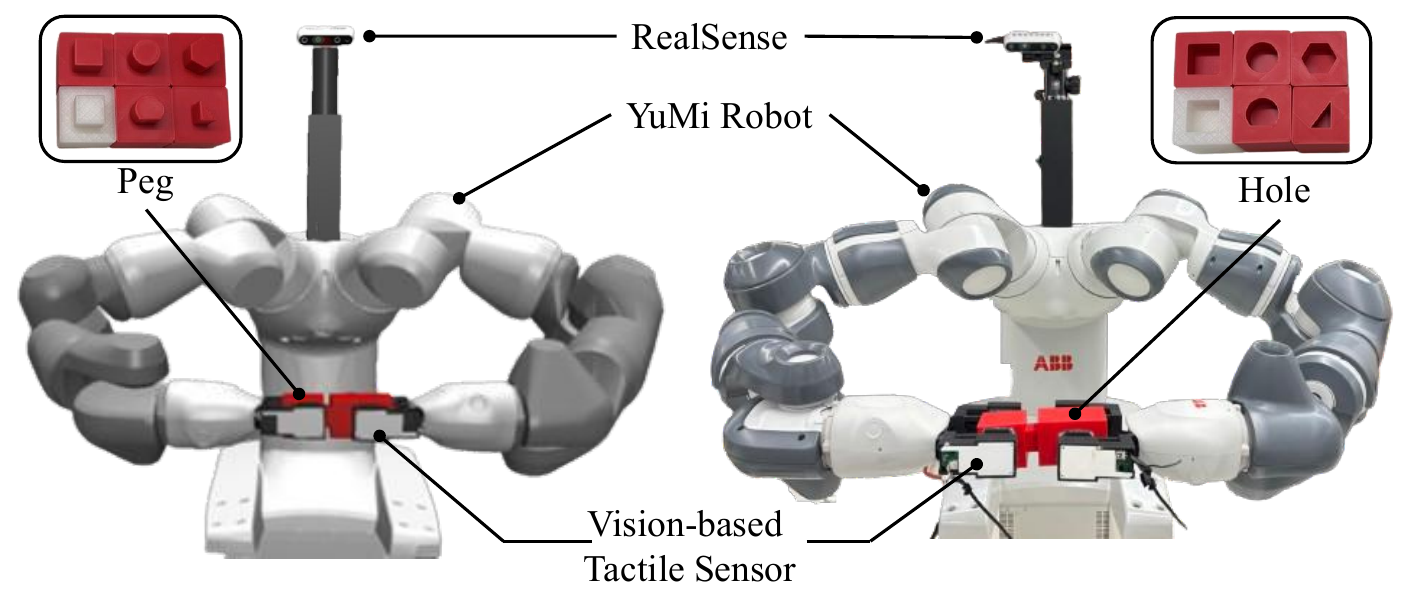}
        \caption{The simulated (left) and real-world (right) experimental setup. We utilize an ABB YuMi robot, equipped four vision-based tactile sensors on its fingers, and mounted an Intel RealSense camera on its body.}
        \label{fig.hardware_setup}
	\vspace{-10pt}
\end{figure}

\section{EXPERIMENTAL RESULTS}
\subsection{Experimental and Training Setups}
We build both simulated and real-world environments for PooH and PiH tasks (Fig.~\ref{fig.hardware_setup}). Both setups use an ABB YuMi dual-arm robot with four GelSight-like vision-based tactile sensors and an Intel RealSense camera. We define each related peg and hole as an object pair; all pairs are 3D-printed (PLA/PETG). Six object pairs are used: red cube, red cylinder, red hexagon, white cube, red D-shape, and red scalene triangle. The clearance is $0.5$\,mm, $1$\,mm, or $2$\,mm. Red cube and red D-shape are seen objects, while the other four are unseen. The simulation replicates the real setup in MuJoCo~\cite{todorov2012mujoco}, using the YuMi model from RoboSuite~\cite{zhu2020robosuite} and the tactile sensor model from~\cite{gomes2021generation}. The simulated motor damping and stiffness are set to 0.1~N$\cdot$m$\cdot$s/rad and 0.1~N$\cdot$m/rad, respectively. We simulate marker motions of the tactile sensor in FOTS~\cite{zhao2024fots}.

For PooH, we train the policy in simulation for $1e5$ steps with $50$ steps per episode.
Simulation is chosen for learning PooH because it enables safe, scalable exploration over diverse contact conditions, yielding broader-coverage reversed demonstrations than what is typically practical to collect in real hardware.
The initial left/right gripper positions in the robot base frame are $(0.5, 0.1, 0.2)$,m and $(0.5, -0.1, 0.2)$,m. Desired goals are sampled in the gripper frame as $(\Delta x,\Delta y,\Delta z)$: for the left gripper, $\Delta x\in(-0.02,0.02)$, $\Delta y\in(0.02,0.04)$, $\Delta z\in(-0.02,0.02)$,m; for the right gripper, $\Delta x\in(-0.02,0.02)$, $\Delta y\in(-0.04,-0.02)$, $\Delta z\in(-0.02,0.02)$,m.

For PiH, we train the policy for $1e5$ episodes using an expert replay buffer built from $500$ simulated and $20$ real-world trajectories per seen object. The object-pair initial positions match the PooH desired goals, and the initial postures are randomized by weights sampled from $[0.8,1.2]$.

\begin{figure}[!t]
	\centering
	\includegraphics[scale=0.45]{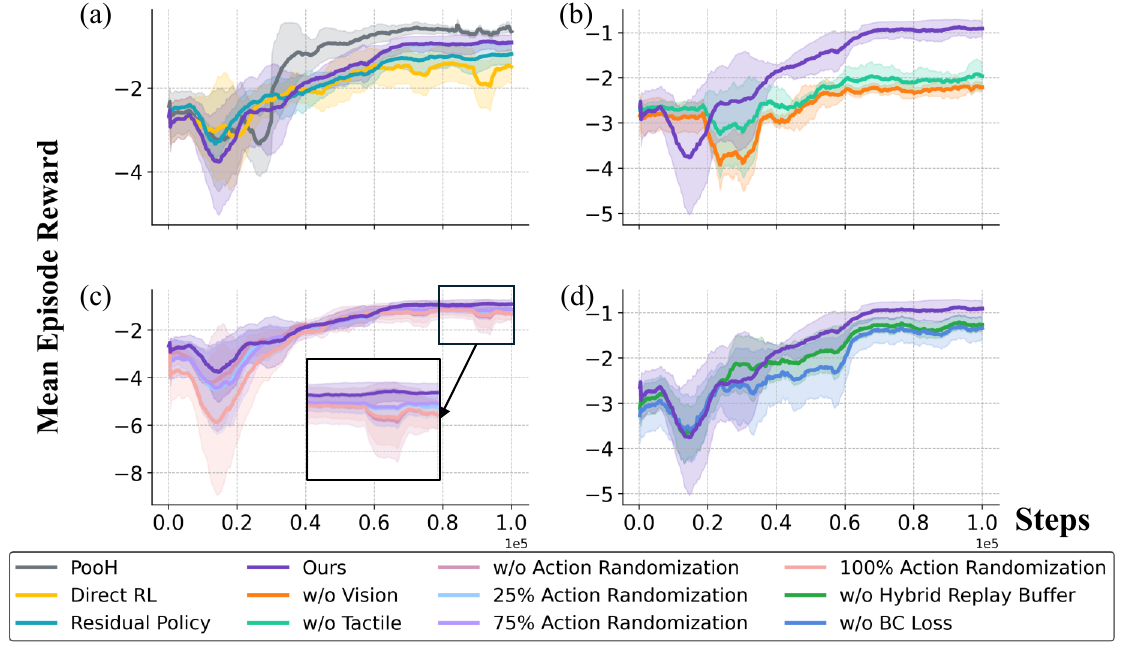}
        \caption{Reward curves for PooH and PiH training variants. (a) Baseline comparison; (b)Ablation on sensing modalities; (c) Ablation on demonstration generation; (d) Ablation on PiH policy learning. Our visual–tactile PooH-to-PiH method (Ours) outperforms direct PiH RL and unimodal ablations, demonstrating the benefit of PooH-derived demonstrations and multimodal sensing. The training random seeds are 0, 25, 50, 75, and 100.}
        \label{fig.reward}
	\vspace{-10pt}
\end{figure}

\begin{figure}[!t]
	\centering
	\includegraphics[scale=0.8]{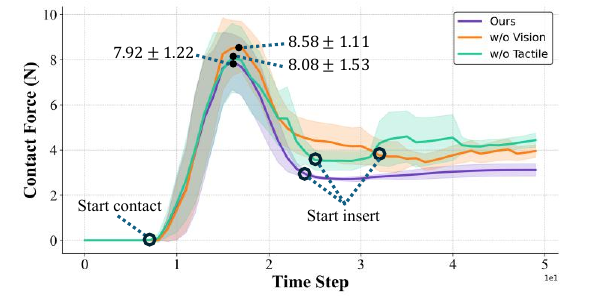}
        \caption{Contact forces during peg-in-hole in simulation. Vision reduces impact at first contact by improving approach accuracy, tactile feedback regulates forces during insertion, and their combination yields the lowest overall force profile.}
        \label{fig.force}
	\vspace{-10pt}
\end{figure}

\subsection{Baseline Comparison}
We compared the proposed method with the following baseline methods:
\begin{itemize}
\item Direct RL: Train the peg-in-hole task from scratch using the same reinforcement learning algorithm (Soft Actor-Critic, SAC) and reward function in Eq. \ref{eq:1}.
\item Supervised Learning (SL): Train the policy via supervised learning on the collected reversed data, with a fully connected linear layer added after the encoders to directly output actions.
\item Residual Policy: Starting from the supervised-learning model described above, we further train the policy with SAC to learn residual end-effector pose corrections.
\end{itemize}

For the Direct RL method, the environment, state space, and action space are identical to those in the PooH-to-PiH setting. The training reward curves of Direct RL and our method are presented in Fig.~\ref{fig.reward} (a), and the success rates of all methods are reported in Fig.~\ref{fig.sim_result}. We evaluated all models on the four objects, with 20 trials per object. The proposed method achieves an average success rate of 80.6\% across all objects, which is 18.1\% higher compared to Direct RL, as well as higher and smoother reward curves. 

The SL model yields a 11.4\% lower average success rate than the proposed method. The residual policy achieves performance comparable to our method on seen objects, but obtains an average success rate that is 4.6\% lower on unseen objects. Although it outperforms the supervised-learning model, its generalization capability remains limited.

In addition, we also demonstrate that PooH tasks have lower complexity than PiH insertion, with higher rewards and faster convergence, as shown in Fig.~\ref{fig.reward} (a). 
These results demonstrate that integrating reversed PooH trajectories into RL training enhances policy performance.

\subsection{Component Ablation Study}
To assess the contribution of individual components, including sensing modalities, the action randomization module, hybrid replay buffers, and behavior cloning (BC) loss, we performed an ablation study. Specifically, we compared the proposed system against variants trained without visual input, without tactile input, without action randomization, without hybrid replay buffers, and without BC loss. Each model was evaluated in 20 trials per object, and the results were aggregated. As shown in Fig.~\ref{fig.reward} (b-c) and Fig.~\ref{fig.sim_result}, both visual and tactile information, along with the three auxiliary components, significantly contribute to overall performance. 

\subsubsection{\textbf{Ablation on Sensing Modalities}}
We ablate the sensing modalities by removing visual or tactile inputs and measuring (1) PiH success rate and (2) peg–hole contact force from MuJoCo, averaged over 10 successful trials. As shown in Fig.~\ref{fig.reward}(b), Fig.~\ref{fig.force}, and Fig.~\ref{fig.sim_result}, removing vision reduces the average success rate to 51.2\% and leads to failures dominated by large approach misalignment. Removing tactile feedback yields an average success rate of 42.3\%, with failures occurring mainly after initial contact due to residual misalignment. \highlight{The large performance degradation without visual input is mainly due to the loss of pre-contact alignment and online error correction. Before contact, tactile feedback is largely unavailable, and proprioception alone cannot fully resolve target pose uncertainty. As a result, small deviations in the initial state and free-motion trajectory can accumulate and cause noticeable misalignment at contact.}

Vision provides global state cues, improving approach accuracy and reducing impact at first contact. In contrast, tactile sensing provides local contact feedback, suppressing excessive forces during insertion. Combining both modalities yields the lowest overall force profile, reducing the maximum contact force by 6.4\% relative to single-modality ablations, indicating that vision guides the robot toward demonstration-consistent contacts while tactile feedback handles misalignment and contact variability from action randomization. We do not include an explicit force penalty in the reward, as it would bias the policy toward conservative, low-contact behaviors and reduce exploration of contact-rich states critical for learning insertion and jamming recovery, despite potentially reducing contact forces.

\subsubsection{\textbf{Ablation on Demonstration Generation}} We also examined the impact of excluding the action randomization module to evaluate its contribution to policy performance. Without action randomization during demonstration generation, the average success rate of the trained PiH model decreases by 14.8\%, especially on unseen objects, indicating that it improves policy robustness. Besides, we tested different ratios (0\%, 25\%, 50\%, 75\%, 100\%) of action randomization. As Fig.~\ref{fig.reward}(c) shows, the reward can reach highest when choosing 50\% action randomization.

\subsubsection{\textbf{Ablation on PiH Policy Learning}}
We study the impact of the hybrid replay buffer and the BC loss on PiH policy learning. Removing either one degrades performance to near the Direct RL baseline, with average success rates of 71.4\% and 65.8\%. This indicates that both are key to exploiting reversed PooH trajectories, improving sample efficiency, and stabilizing training. We also do not expect the auxiliary BC loss to reduce generalization: the demonstrations are diverse and cover the PiH distribution, while BC is lightly weighted and annealed, stabilizing early training and letting RL dominate later. In addition, we do not include a Q-filtered BC term, as it may constrain policy exploration, despite potentially stabilizing training.

\begin{figure}[!t]
	\centering
	\includegraphics[scale=0.43]{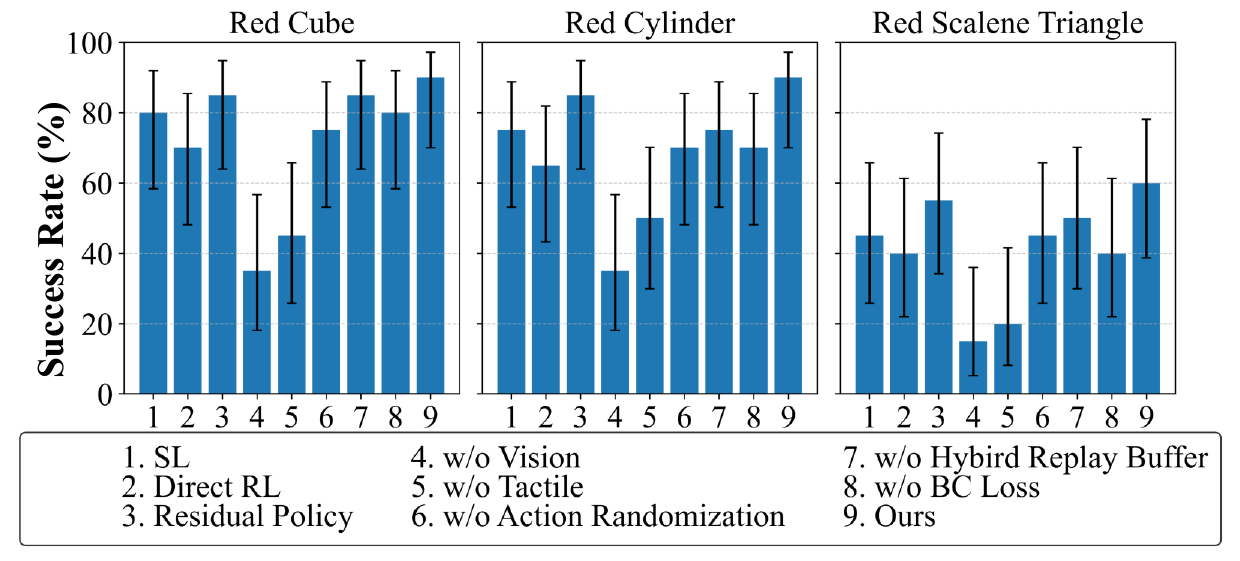}
        \caption{Peg-in-hole success rates (95\% Wilson CIs) compared with other methods under 1.0 mm clearance. The results of all other objects are available on \href{https://sites.google.com/view/pooh2pih}{our project website}.}
        \label{fig.sim_result}
	\vspace{-10pt}
\end{figure}

The results (20 trials per object; Fig.~\ref{fig.sim_result}) show that the proposed vision–tactile method effectively learns PiH from reversed PooH trajectories; ablations further confirm that both the PooH-to-PiH data pipeline and multimodal observations are essential, as removing either degrades performance toward Direct RL.


\begin{table}[t]
\centering
\caption{Sim-to-Real n successes / N of the proposed method}
\label{Tab:2}
\setlength{\tabcolsep}{6pt}
\renewcommand{\arraystretch}{0.6}
{\small
\begin{tabular}{ccccc}
\toprule
Material & \makecell{Clearance\\(mm)} & \makecell{Red\\Cube} & \makecell{Red\\Cylinder} & \makecell{Red Scalene\\Triangle} \\
\midrule
\multirow{3}{*}{PLA}
 & 0.5 & 14/20 & 15/20 & 8/20 \\
 & 1.0 & 18/20 & 17/20 & 11/20 \\
 & 2.0 & 19/20 & 20/20 & 14/20 \\
\midrule
\multirow{3}{*}{PETG}
 & 0.5 & 13/20 & 13/20 & 6/20 \\
 & 1.0 & 16/20 & 16/20 & 10/20 \\
 & 2.0 & 19/20 & 18/20 & 14/20 \\
\bottomrule
\end{tabular}
\vspace{-10pt}
}
\end{table}

\subsection{Sim-to-Real Policy Transfer}
\begin{figure}[!t]
	\centering
	\includegraphics[scale=0.35]{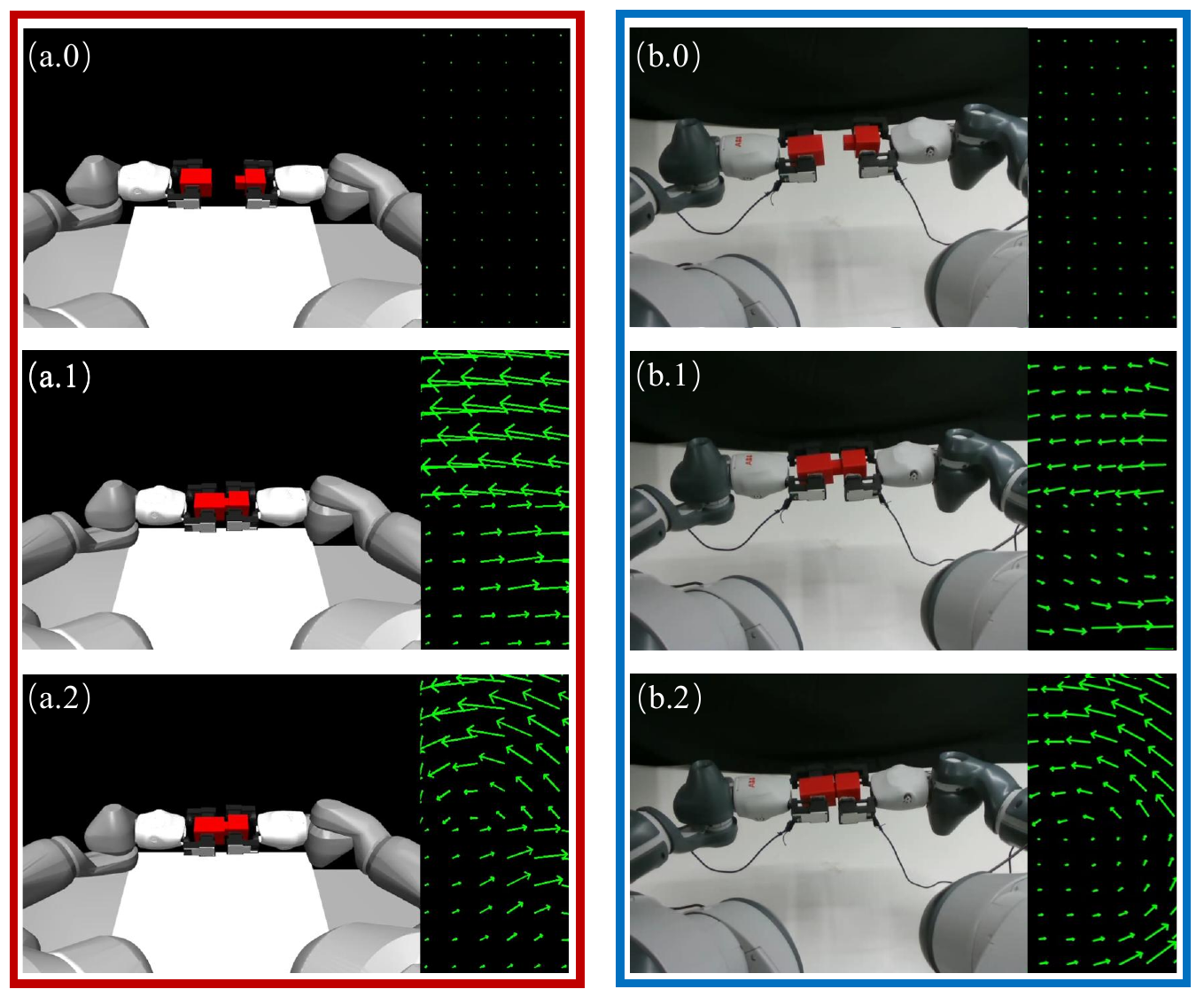}
        \caption{Simulated and real-world peg-in-hole experiments with cube objects. Each snapshot shows the robot-side visual image (left) and tactile images from the left and right grippers (right). (a.0–2) Simulation; (b.0–2) Real world. The learned visual–tactile PiH policy achieves similar alignment behavior and observation patterns in both domains.}
        \label{fig.real_world_exp}
        \vspace{-10pt}
\end{figure}

After training the PiH policies in simulation, we transfer them to real-world execution. Real-world deployment uses the same observation space as in simulation and does not rely on any privileged inputs. To reduce the sim-to-real gap and enhance transfer performance, we consider from the following three aspects:

\subsubsection{\textbf{Tactile Calibration}}

We use a data-driven tactile simulator adapted from~\cite{zhao2024fots} to generate high-fidelity tactile flows calibrated with real tactile images.
Tactile flow data are collected from the calibrated simulator using the calibration objects from~\cite{gomes2021generation}. For each object, 100 frames are recorded with randomized relative translations in [-1.0, -1.0, -0.05]–[1.0, 1.0, 0.5] cm and rotations in [-0.2, -0.2, -0.5]–[0.2, 0.2, 0.5] rad, resulting in 2100 frames in total. PCA is then applied to the collected data to obtain compact and robust representations~\cite{xue2025reactive}.
Using a 15D PCA representation together with well-calibrated coefficients is important for maintaining PiH performance.
However, a residual sim-to-real gap in marker motion remains due to calibration with a single sphere indenter and the simplification to three primitive motions (dilate, shear, and twist).

\subsubsection{\textbf{Visual Domain Randomization}}
We apply domain randomization by varying object colors in the simulated environment. Specifically, the RGB values of simulated objects are multiplied by random scaling factors within the range [0.8, 1.2], effectively improving generalization to real-world lighting and material conditions. However, there is still a gap in the visual observations between the simulation and the real world, e.g., cables and connectors in real world.

\subsubsection{\textbf{Real-World Demonstration Collection}}
To better align simulation with the real world, 20 real-world PooH demonstrations are collected for each known object and then reversed into PiH trajectories. These real-world data are mixed with simulated data in a hybrid replay buffer to train the PiH policy.

During deployment, the initial gripper poses are randomized within a predefined range (Fig.~\ref{fig.real_world_exp}(b.0)). The robot executes the learned visual-tactile policy for alignment and insertion (Fig.~\ref{fig.real_world_exp}(b.0–2)). We test six object pairs with 20 trials each; Specifically, a success case is regarded by measuring insertion depth, which should exceed 1cm.
Partial results are summarized in Table~\ref{Tab:2} (Additional results and videos are available on \href{https://sites.google.com/view/pooh2pih}{our project website}.). Although performance slightly drops from simulation, our method achieves a 72.1\% average success rate, demonstrating effective sim-to-real transfer and generalization across diverse geometries.

\section{CONCLUSIONS}
In this work, a visual-tactile skill learning framework is proposed for peg-in-hole assembly to enable efficient insertion learning. The key idea is to learn a visual-tactile PooH policy and reuse its temporally inverted and action-randomized trajectories as demonstrations for PiH learning within the same POMDP. Visual and tactile sensing provide robust approach cues and contact feedback, as validated by the ablation studies. The proposed method outperforms supervised learning and direct RL baselines, and is validated in both simulation and real-world experiments across diverse peg and hole shapes. \highlight{Future work will further reduce the sim-to-real gap by improving tactile simulator fidelity, applying richer visual adaptation or fine-tuning with real-world data, and narrowing residual dynamics/contact mismatches.} The framework will also be extended to manipulation tasks that can benefit from their reverse tasks.

\bibliographystyle{IEEEtran}
\bibliography{reference}

\end{document}